\documentclass[runningheads]{llncs}
\usepackage[T1]{fontenc}
\usepackage{graphicx}
\usepackage{booktabs}
\usepackage[misc]{ifsym}
\newcommand{\corr}{(\Letter)}
\newcommand{\defeq}{\overset{\text{\tiny def}}{=}}
\usepackage{comment}
\usepackage{xcolor}
\usepackage{mwe}
\usepackage{amsmath} 
\usepackage{amssymb} 
\usepackage{multirow} 
\usepackage{bbding} 
\usepackage{float}
\usepackage{blindtext}
\usepackage{hyperref}
\usepackage{algorithm}
\usepackage{algorithmic}

\begin{document}

\newcommand{\karim}[1]{{\color{red}#1} }
\def\etc{etc.\@\xspace}
\def\ie{\textit{i.e.}\@\xspace}
\def\eg{\textit{e.g.}\@\xspace}
\def\R{\mathbb{R}}
\def\E{\mathbb{E}}
\def\x{\mathbf{x}}
\def\g{\mathbf{g}}
\def\N{\mathbf{N}}
\def\G{\mathbf{G}}
\def\U{\mathbf{U}}
\title{KCLNet: Physics-Informed Power Flow Prediction via Constraints Projections}

\titlerunning{KCLNet: Integrating Physics Knowledge via Constraints Projections}


\author{Pantelis Dogoulis \orcidID{0000-0003-3843-9927} \corr  \and
Karim Tit \orcidID{0009-0000-5394-3078} \and
Maxime Cordy \orcidID{0000-0001-8312-1358}}

\authorrunning{P. Dogoulis et al.}


\institute{SerVal, SnT, University of Luxembourg 
\\
\email{\{panteleimon.dogoulis,karim.tit,maxime.cordy\}@uni.lu}
}

\tocauthor{Pantelis Dogoulis,Karim Tit, Maxime Cordy}
\toctitle{KCLNet: Integrating Physics Knowledge via Constraints Projections}

\maketitle  

\begin{abstract}

In the modern context of power systems, rapid, scalable, and physically plausible power flow predictions are essential for ensuring the grid’s safe and efficient operation. While traditional numerical methods have proven robust, they require extensive computation to maintain physical fidelity under dynamic or contingency conditions. In contrast, recent advancements in artificial intelligence (AI) have significantly improved computational speed; however, they often fail to enforce fundamental physical laws during real-world contingencies, resulting in physically implausible predictions. In this work, we introduce KCLNet, a physics-informed graph neural network that incorporates Kirchhoff's Current Law as a hard constraint via hyperplane projections. KCLNet attains competitive prediction accuracy while ensuring zero KCL violations, thereby delivering reliable and physically consistent power flow predictions critical to secure the operation of modern smart grids.

\end{abstract}

\section{Introduction}
\label{sec:intro}
The evolution of modern power systems has introduced novel challenges for reliable and efficient grid operation. At the heart of these challenges lies the power flow prediction problem, a fundamental task that involves computing the steady-state operating conditions of a power grid under varying load, generation, and contingency conditions. Traditionally, this problem has been solved using numerical methods that solve non-linear equations derived from Kirchhoff’s laws \cite{akram2015newton,sereeter2019comparison,wasley1974newton}. However, the increasing complexity and scale of power systems, driven by the integration of renewable energy sources and distributed generation, have required the development of more scalable and adaptive techniques \cite{leyli2022lips}.

Recent advances in machine learning \cite{lin2024powerflownet,kody2022modeling,donon2020leap} offer a promising alternative by leveraging simulated and real-time data to predict power flow dynamics with speed and accuracy. However, the power flow prediction problem is inherently related to the physical laws that govern electrical networks. Inaccurate predictions can have severe consequences, particularly under contingency conditions, where the failure of a single component must not compromise system stability. This is known as the N-1 criterion and is a widely recognized reliability standard in power grid operations \cite{zima2005security,dogoulis2024robustness}. Consequently, there is a critical need for machine learning models that are capable of producing physically plausible solutions, even in real-world contingency scenarios.

Physics-Informed Machine Learning (PIML) \cite{karniadakis2021physics,qian2020lift} has emerged as a robust theoretical framework to address this challenge by integrating domain-specific physical principles directly into the learning process. 
Conventional PIML strategies typically fall into one of three categories: (a) incorporating soft constraints via the loss function to penalize deviations from physical laws; (b) leveraging simulation results or specialized weight initialization as a foundation for model training; or (c) embedding the relevant physical principles directly into the model architecture \cite{huang2022applications}. In the context of power flow prediction, most of the previous works have predominantly relied on the soft constraint approach. 

In this work, we focus on the integration of Kirchhoff's Current Law, a cornerstone of electrical circuit theory, directly into the model, as a hard linear constraint. Even though, under some strong theoretical assumptions, the conventional approach of imposing soft constraints in the loss function can guide the learning process, it does not guarantee absolute adherence to physical laws. This limitation is increasingly prominent in extreme or unseen operating conditions like the N-1 scenario. By contrast, our approach enforces Kirchhoff’s law as a strict equality constraint within the optimization problem, ensuring that every prediction generated by the model is physically plausible \footnote{\footnotesize{Code Repo: \url{https://github.com/dogoulis/ecml-pf-pred}}}. This is especially crucial for power flow prediction, where even minor deviations from established laws can propagate into significant operational risks under contingency scenarios.

Our contribution can be summarized as follows:
\begin{itemize}
    \item We propose a novel physics-informed machine learning model that integrates Kirchhoff’s law as a hard equality constraint, ensuring physically accurate predictions under normal operating conditions and contingency scenarios.
    \item We provide a critical assessment of the limitations inherent to conventional soft constraint approaches, particularly in the context of N-1 contingency scenarios.
    \item We validate our methodology using simulated data depicting real-world operating grid scenarios, demonstrating enhanced reliability and operational safety in modern power systems.
\end{itemize}

\section{Definition of the Power Flow Prediction Problem}
\label{sec:form}

The power flow prediction problem is central to the analysis and operation of electrical power systems, as it involves determining the steady-state conditions of a grid based on prescribed inputs such as generation levels, load demands, and network topology. Conventionally, the problem is formulated as a set of nonlinear algebraic equations derived from Kirchhoff's Current Law (KCL), which are typically solved using iterative numerical methods like the Newton-Raphson algorithm. In this section, we first review the classical formulation and solution methodologies for the power flow problem before discussing its reformulation within a machine learning framework. For clarity, we begin with a concise overview of the power grid terminology and we briefly describe one important security criterion for the operation of the power grids.

\paragraph{Power Grid Terminology:}
In power system analysis, a \emph{bus} is a node where it represents elements such as generators, loads, or transformers and is generally characterized by four parameters: active power ($P$), reactive power ($Q$), voltage magnitude ($V_m$) and phase angle ($V_a$). Buses can be categorized into generators (\textit{PV} buses), loads (\textit{PQ} buses) or slack buses. The latest serves as the grid's reference point by absorbing active and reactive power, based on the discrepancies of the other nodes. A \emph{transmission line} is a branch that links two buses, with its behavior described by electrical parameters such as resistance ($r$) and reactance ($x$).

\paragraph{N-1 criterion:} In power grids, the N-1 criterion is a reliability standard that mandates the grid must continue to operate securely even if any single component (such as a transmission line) fails. This criterion ensures that the network has sufficient redundancy and reserve capacity, so that the loss of one element does not lead to system instability or compromise the continuity of the power supply.

\paragraph{Kirchhoff's Current Law:}
At the core of power system analysis lie Kirchhoff's laws, which govern the conservation of energy and charge in electrical networks.
KCL states that the algebraic sum of currents entering a node must equal the sum of currents leaving the node. In power grids which we are interested in, this principle can be expressed in terms of the active and the reactive power, under some operational hypotheses. At any bus \(i\), the net active power injection \(P_i\) is equal to the sum of the active powers \(P_{ij}\) flowing through all the transmission lines connected to that bus:
\begin{equation}
\label{eq:kcl-bus}
\begin{cases}
    P_i = \sum_{j \in \mathcal{N}(i)} P_{ij}
   \\
   Q_i = \sum_{j \in \mathcal{N}(i)} Q_{ij}
\end{cases}
\end{equation}
where $P_i, Q_i$ are the active and reactive powers injected at bus $i$, $P_{ij}, Q_{ij}$ represent the active and the reactive powers flowing from bus $i$ to bus $j$, and
$\mathcal{N}(i)$ denotes the set of buses connected to bus $i$.
Using Kirchhoff's laws as a foundation, we can derive a set of nonlinear algebraic equations which describe the power balance at each bus. These equations incorporate other physical entities of the grid (voltage magnitudes and angles) as well as the transmission line's intrinsic characteristics. For an \(N\)-bus system, active (\(P_i\)) and reactive (\(Q_i\)) power injections at bus \(i\) are given by:

\begin{equation}
\begin{cases}
P_i = V_i \sum_{j=1}^{N} V_j \left( G_{ij} \cos\theta_{ij} + B_{ij} \sin\theta_{ij} \right) 
\\
Q_i = V_i \sum_{j=1}^{N} V_j \left( G_{ij} \sin\theta_{ij} - B_{ij} \cos\theta_{ij} \right) \end{cases}
\label{eq:KCL-ij}
\end{equation}

where $V_i$ is the voltage magnitude at bus $i$, $\theta_{ij} = \theta_i - \theta_j$ is the phase angle difference between buses $i$ and $j$, while $G_{ij}$ and $B_{ij}$ denote the conductance and susceptance of the transmission line connecting $i$ and $j$.
The resulting system of nonlinear equations is commonly solved using iterative numerical techniques. One of the most widely used methods is the \textit{Newton-Raphson} method, which is favored for its quadratic convergence properties. The method iteratively refines an initial guess of the powers and the voltages by solving the following update equation:

\begin{equation}
\mathbf{x}^{(k+1)} = \mathbf{x}^{(k)} - \mathbf{J}^{-1}(\mathbf{x}^{(k)}) \, \mathbf{f}(\mathbf{x}^{(k)}),
\end{equation}

where: $\mathbf{x}$ is the vector of state variables, $\mathbf{f}(\mathbf{x})$ represents the mismatch between the calculated and specified power injections, $\mathbf{J}(\mathbf{x})$ is the Jacobian matrix of partial derivatives of $\mathbf{f}$ with respect to $\mathbf{x}$.
Although the Newton-Raphson method and its variants provide reliable solutions under normal operating conditions, they can be computationally intensive for large-scale networks or in scenarios with rapidly changing system conditions (i.e. contingencies).

\section{Related Work}
\label{sec:lit-rev}
Recent advances in applying artificial intelligence to power flow prediction can be broadly divided into methods that rely primarily on learning from data without explicit physical constraints, and methods that incorporate physics-based guidance or constraints during training. An example of the former is described in \cite{donnot2018fast}, where the authors propose a neural network to tackle the power flow prediction problem based on a guided-dropout technique, and they validate their approach in different contingency cases. Similarly, the authors in \cite{donon2020leap} propose LeapNet, a latent-space network that handles structural or parametric shifts by encoding inputs into a hidden representation, applying a conditional “leap” dependent on discrete topology changes, and decoding back to flow predictions. Furthermore, in \cite{liu2018data}, the authors introduce two modeling approaches: a partial least squares (PLS) regression model and a Bayesian linear regression model. Both models are derived from a linearized formulation of the underlying problem, which is obtained by adopting specific assumptions regarding grid operations.

In contrast, several \textit{physics-informed} approaches add explicit power-flow constraints. The authors in \cite{kody2022modeling} propose a method that enforces piecewise-linearity, combining a baseline linearization (derived from the Jacobian at a nominal operating point of the grid) with ReLU-based low-rank updates to compactly approximate the power flow equations, thus introducing a physics-based term into the network's architecture. PowerflowNet \cite{lin2024powerflownet} employs a graph neural network architecture equipped with message passing layers, augmented by an additional loss term representing the squared mismatch between the neural network's predicted power injections and the injections computed from the power-flow equations, thus penalizing physical inconsistency. The authors in \cite{de2022physics} propose a physics-informed geometric deep learning scheme that encodes the grid's topology via graph neural networks, with partial derivatives or power flow constraints contributing to the training loss and encouraging alignment with Kirchhoff’s laws. Finally, the authors at \cite{donon2020neural} present a Graph Neural Solver for AC power flows that leverages message passing across nodes and edges while penalizing Kirchhoff’s law violations. Most of these works can be broadly classified under the term of PIML, where physical constraints are included into the loss function as an additional term. Concretely, the total loss function can be formulated as follows:

\begin{equation}
    \mathcal{L}(\theta) 
    = \underbrace{\mathcal{L}_{\text{reg}}(\theta)}_{\text{MSE}} 
    \;+\; \mathcal{L}_{\text{physics}}(\theta),
\end{equation}

where $\mathcal{L}_{\text{reg}}(\theta)$ is a standard regression loss (commonly the mean squared error) that measures the discrepancy between the model’s predictions and the ground-truth data, and $\mathcal{L}_{\text{physics}}(\theta)$ is a term enforcing physical consistency, most commonly by penalizing the mismatch in the power flow equations described in \ref{eq:KCL-ij}. Additionally, it is important to note that the formulation of power flow prediction is not universally standardized; some approaches focus on estimating transmission line flows while others predict nodal injections. However, by leveraging the governing physical equations, these formulations can often be interchanged, although the extent of this conversion is inherently dependent on the initial problem definition. In our approach, crucially, physical constraints are embedded \emph{directly} into the model architecture, thereby ensuring that all predictions rigorously conform to the underlying physical laws (see Section \ref{sec:method} below).

\section{Proposed Method}
\label{sec:method}
\begin{figure*}[h]
  \centering
  \includegraphics[width=0.75\textwidth]{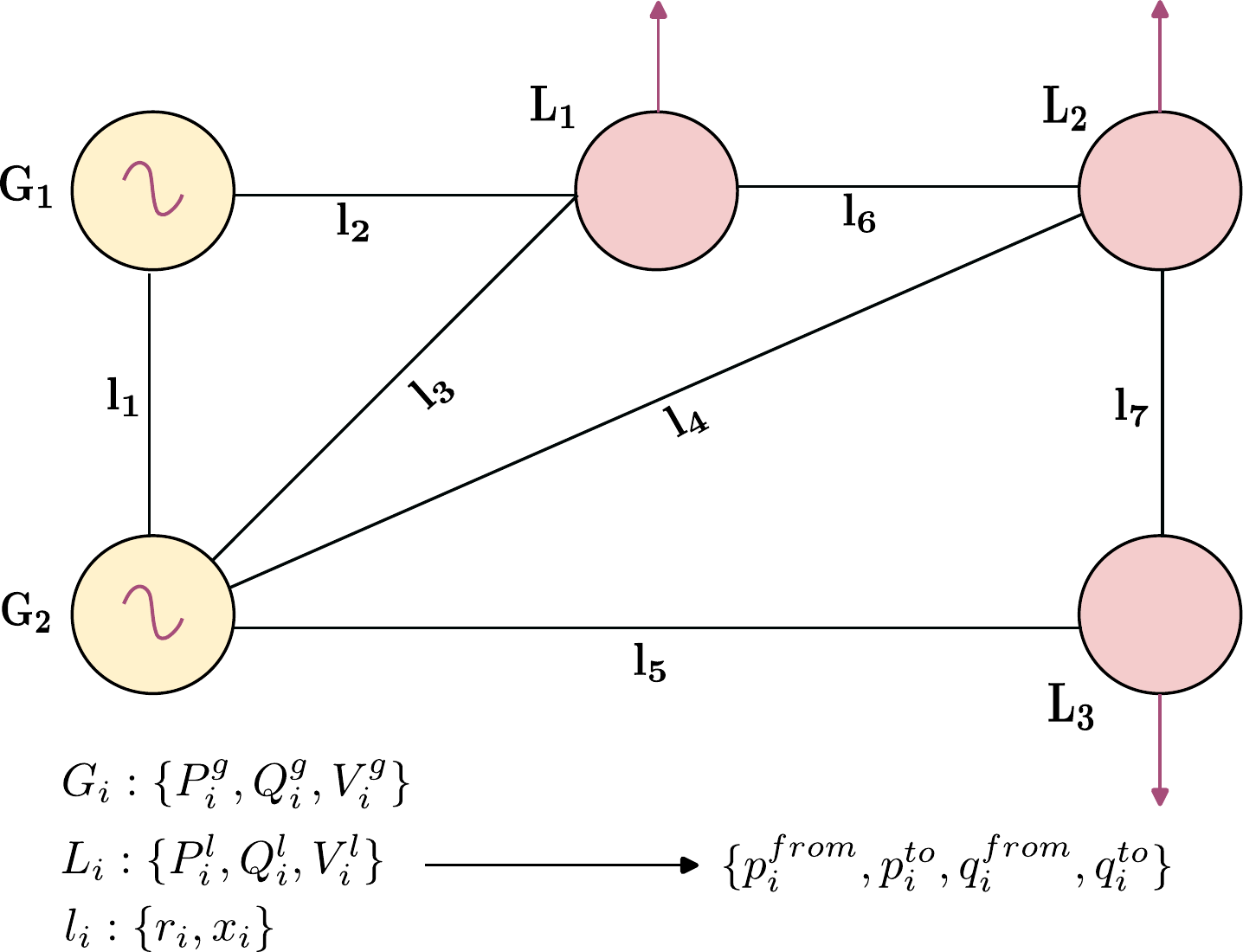}
  \caption{Illustration of a power grid (IEEE5) represented as a graph. Two generator nodes ($\mathbf{G_1}$ and $\mathbf{G_2}$, shown in yellow) are connected to three load nodes ($\mathbf{L_1}$, $\mathbf{L_2}$, and $\mathbf{L_3}$, shown in pink) through multiple transmission lines (labeled $\mathbf{l_1}$, $\mathbf{l_2}$, $\mathbf{l_3}$, $\mathbf{l_4}$, $\mathbf{l_5}$, $\mathbf{l_6}$, and $\mathbf{l_7}$). Each generator node is associated with the node features $\{P_i^g, Q_i^g, V_i^g\}$, while each load node is associated with $\{P_i^l, Q_i^l, V_i^l\}$. The edge features are ($\{r_i, x_i\}$), which represent the intrinsic electrical characteristics of the transmission lines (as discussed in Section \ref{sec:form}). The formulation of the problem is to predict the associated from- and to-bus vector: ($p_i^{\mathrm{from}}, p_i^{\mathrm{to}}, q_i^{\mathrm{from}}, q_i^{\mathrm{to}}$).
  }
  \label{fig:grid-rep}
\end{figure*}

\subsection{Grid Represented as a Graph}
A power grid can be naturally modeled as a graph \(\mathcal{G} = (\mathcal{V}, \mathcal{E})\), where the set of nodes \(\mathcal{V}\) represents buses and the set of edges \(\mathcal{E}\) corresponds to the transmission lines connecting them (see Figure \ref{fig:grid-rep}). Each node is endowed with three attributes: active (${P_i}$) and reactive power injections (${Q_i}$) and voltage magnitudes (${V_i}$) (note that here we omit the $V_a$ feature), while each edge is characterized by intrinsic electrical parameters which are the resistance (${r_i}$) and the reactance (${x_i}$). 
The principal objective is to compute the active and reactive power flows at both ends of each edge (transmission line).

\begin{figure*}[htbp]
  \centering
  \includegraphics[width=\textwidth]{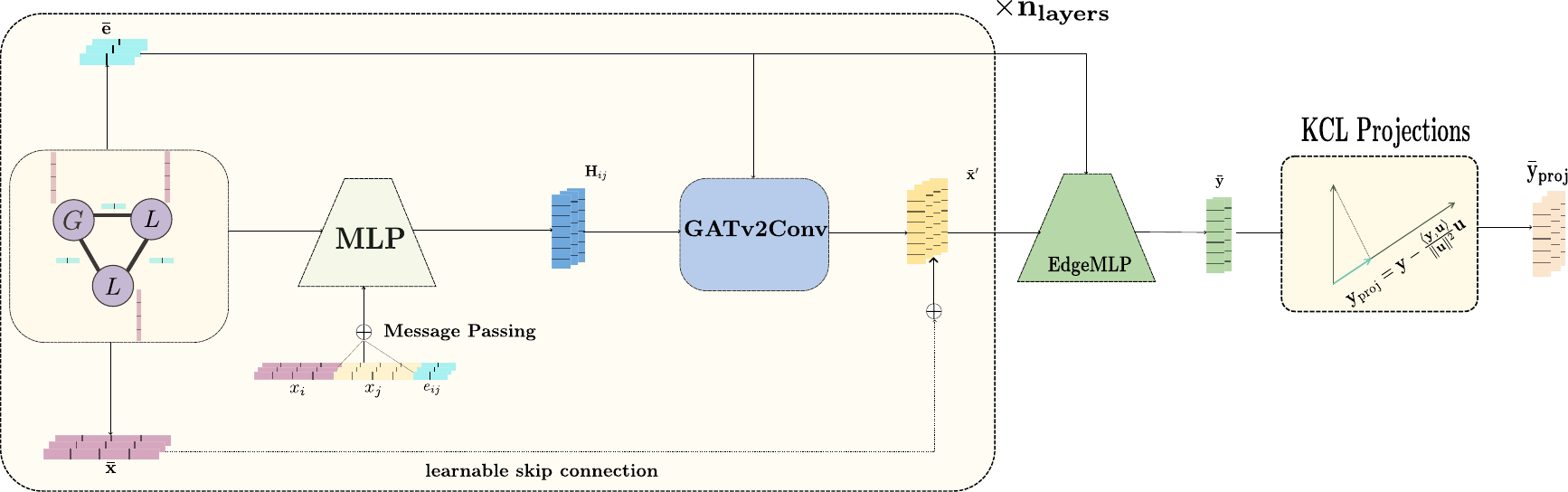}
  \caption{Illustration of the proposed KCLNet.}
  \label{fig:method}
\end{figure*}

\subsection{Model Architecture}
Figure~\ref{fig:method} presents an overview of the complete architecture of the proposed GNN-based model for power flow prediction. The model takes as input the node features \(\mathbf{x}_i\) and the edge attributes \(\mathbf{e}_{ij}\), which together encode the electrical and topological characteristics of the network. Mathematically, the overall model implements a function
\[
f: (\mathbf{X}, \mathbf{E}) \rightarrow \hat{\mathbf{Y}},
\]
where \(\mathbf{X} \in \mathbb{R}^{N \times 3}\) is the node feature matrix, \(\mathbf{E} \in \mathbb{R}^{|E| \times 2}\) is the edge attribute matrix, and \(\hat{\mathbf{Y}} \in \mathbb{R}^{|E| \times 4}\) contains the predicted flow parameters for each edge.
In the initial stage, an MLP-based message passing module inspired by \cite{lin2024powerflownet} processes the concatenated vector
\[
\mathbf{z}_{ij} = \left[\mathbf{x}_i,\, \mathbf{x}_j,\, \mathbf{e}_{ij}\right],
\]
for each edge \((i,j)\). This module computes the hidden state as:
\[
\mathbf{h}_{ij} = W_2\, \text{LeakyReLU}\big(W_1\, \mathbf{z}_{ij} + \mathbf{b}_1\big) + \mathbf{b}_2,
\]
where \(W_1\) and \(W_2\) are weight matrices and \(\mathbf{b}_1\) and \(\mathbf{b}_2\) are bias vectors. The intermediate embedding for node \(i\) is then obtained by aggregating the messages from its neighbors:
\[
\mathbf{x}_i' = \sum_{j \in \mathcal{N}(i)} \mathbf{h}_{ij}.
\]
These intermediate embeddings are further refined by a $\mathbf{GATv2Conv}$ layer \cite{brody2021attentive}, which employs a multi-head attention mechanism. In this layer, both the updated node embeddings \(\mathbf{x}_i'\) and the edge attributes \(\mathbf{e}_{ij}\) contribute to the computation of attention coefficients \(a_{ij}\), and the refined node embeddings are given by
\[
\mathbf{x}_i'' = \sum_{j \in \mathcal{N}(i)} a_{ij}\, \mathbf{x}_j'.
\]
To preserve the original node information and facilitate effective gradient propagation, a learnable skip connection is incorporated. The original node features are first projected into the hidden space via
\[
\tilde{\mathbf{x}}_i = W_{\text{skip}}\, \mathbf{x}_i,
\]
and then combined with the refined embeddings:
\[
\hat{\mathbf{x}}_i = \mathbf{x}_i'' + \tilde{\mathbf{x}}_i.
\]
Finally, edge-level predictions are derived by constructing representations for each edge. For a given edge \((i,j)\), the final node embeddings corresponding to nodes \(i\) and \(j\) are concatenated with the edge attributes:
\[
\mathbf{z}_{ij}^{\text{edge}} = \left[\hat{\mathbf{x}}_i,\, \hat{\mathbf{x}}_j,\, \mathbf{e}_{ij}\right],
\]
and the resulting vector is passed through an additional MLP, referred to as $\mathbf{EdgeMLP}$, to produce the predicted flow parameters:
\[
\hat{\mathbf{y}}_{ij} = \text{EdgeMLP}\left(\mathbf{z}_{ij}^{\text{edge}}\right).
\]

\subsection{KCL Projections for Physical Constraints}
\label{sub:proj}

Kirchhoff’s Current Law (KCL) states that, at each bus, the total incoming power must equal the total outgoing power (see eq. \eqref{eq:kcl-bus}). To impose this principle, we proceed with the following notations:

\begin{itemize}
    \item For each bus \(i \in \{1,\ldots,N\}\), let \(P_{\text{net}}(i)\) and \(Q_{\text{net}}(i)\) be the \emph{measured} active and reactive power injections, respectively.
    \item For each line (edge) \(e\), let \(p^{\text{from}}_e, p^{\text{to}}_e\) denote the \emph{predicted} active power flows, and \(q^{\text{from}}_e, q^{\text{to}}_e\) the \emph{predicted} reactive power flows.
\end{itemize}

Given a bus \(i\), define its \emph{calculated} active and reactive power injections as
\[
P_{\text{calc}}(i) 
\;=\;
\sum_{e \in \mathcal{E}^{\text{from}}(i)} p^{\text{from}}_e 
\;+\;
\sum_{e \in \mathcal{E}^{\text{to}}(i)} p^{\text{to}}_e,
\quad
Q_{\text{calc}}(i) 
\;=\;
\sum_{e \in \mathcal{E}^{\text{from}}(i)} q^{\text{from}}_e 
\;+\;
\sum_{e \in \mathcal{E}^{\text{to}}(i)} q^{\text{to}}_e,
\]
where \(\mathcal{E}^{\text{from}}(i)\) and \(\mathcal{E}^{\text{to}}(i)\) denote the sets of edges for which bus \(i\) is respectively the source and the target.

\medskip
\noindent
\textbf{KCL in Terms of Predicted Flows.}
From eq.~\eqref{eq:kcl-bus}, the KCL requirement at each bus \(i\) can be expressed as
\begin{equation}
\label{eq:KCL1}
P_\text{net}(i) + P_{\text{calc}}(i)
\;=\;
P_\text{net}(i)
\;+\;
\sum_{e \in \mathcal{E}^{\text{from}}(i)} p^{\text{from}}_e
\;+\;
\sum_{e \in \mathcal{E}^{\text{to}}(i)} p^{\text{to}}_e
\;=\;
0,
\end{equation}
for active power, and
\begin{equation}
\label{eq:KCL2}
Q_\text{net}(i) + Q_{\text{calc}}(i)
\;=\;
Q_\text{net}(i)
\;+\;
\sum_{e \in \mathcal{E}^{\text{from}}(i)} q^{\text{from}}_e
\;+\;
\sum_{e \in \mathcal{E}^{\text{to}}(i)} q^{\text{to}}_e
\;=\;
0,
\end{equation}
for reactive power.

\medskip
\noindent
\textbf{Projection Operators.}
We enforce these constraints by constructing hyperplane projections for each node. Specifically, for bus~\(i\), we define two projections: \(\Pi_i^{P}\) (for active power) and \(\Pi_i^{Q}\) (for reactive power). Both operators act on the prediction vector 
\(\mathbf{y} \in \mathbb{R}^{|E|\times 4}\) to ensure that KCL holds at \(i\).

\paragraph{Normal Vectors.}
First, define the vectors \(\mathbf{a}^{P,i}, \mathbf{a}^{Q,i} \in \mathbb{R}^{|E|\times 4}\) that capture the contribution of each component of \(\mathbf{y}\) to the net power injection at bus~\(i\). From eqs.~\eqref{eq:KCL1} and~\eqref{eq:KCL2}, the entries of \(\mathbf{a}^{P,i}\) are
\[
a^{P,i}_{e,l}
\;=\;
\begin{cases}
1, & \text{if } e \in \mathcal{E}^{\text{from}}(i)\,\text{and}\, l=1,\\
1, & \text{if } e \in \mathcal{E}^{\text{to}}(i)\,\text{and}\, l=2,\\
0, & \text{otherwise},
\end{cases}
\]
and \(\mathbf{a}^{Q,i}\) is defined similarly for reactive power.

\paragraph{Projection Formulas.}
Using these normal vectors, the orthogonal (affine) hyperplane projections that zero out the KCL violations at bus~\(i\) are
\[
\Pi_i^{P}(\mathbf{y})
\;\defeq\;
\mathbf{y}
\;-\;
\frac{\langle \mathbf{a}^{P,i}, \mathbf{y} \rangle+P_{\text{net}}(i)}{\|\mathbf{a}^{P,i}\|^2}
\,\mathbf{a}^{P,i},
\quad
\Pi_i^{Q}(\mathbf{y})
\;\defeq\;
\mathbf{y}
\;-\;
\frac{\langle \mathbf{a}^{Q,i}, \mathbf{y} \rangle+Q_{\text{net}}(i)}{\|\mathbf{a}^{Q,i}\|^2}
\,\mathbf{a}^{Q,i}.
\]
Applying \(\Pi_i^{P}\) removes the component of \(\mathbf{y}\) collinear to $\mathbf{a}^{P,i}$ that violates eq.~\eqref{eq:KCL1}, while \(\Pi_i^{Q}\) does the same for eq.~\eqref{eq:KCL2}.

\medskip
\noindent
\textbf{Sequential Node Projections.}
Combining these local projections for each of the \(N\) buses, and choosing some variable-ordering permutation $\sigma : \{1,\ldots,N\} \to \{1,\ldots,N\}$, we define
\[
\Pi^{\text{KCL}}_{\sigma}
\;\defeq\;
\Pi_{\sigma(1)}^{P} \circ \Pi_{\sigma(1)}^{Q}
\;\circ\;
\Pi_{\sigma(2)}^{P} \circ \Pi_{\sigma(2)}^{Q}
\;\circ\;\cdots\circ\;
\Pi_{\sigma(N)}^{P} \circ \Pi_{\sigma(N)}^{Q}.
\]
Although applying $\Pi^{\text{KCL}}_{\sigma}$ once may not fully enforce all KCL constraints
(since projections generally do \emph{not} commute),
\emph{repeated} application of $\Pi^{\text{KCL}}_{\sigma}$ will asymptotically converge
to a KCL-feasible solution, as in Kaczmarz's iterative hyperplane method~\cite{kaczmarz}.
In practice, however, finding a good ordering $\sigma$ is nontrivial; one often resorts to random selection~\cite{random_kaczmarz}
and convergence can be slow.

\noindent
\subsubsection*{Alternative Formulation.}

An alternative to applying node-by-node projections is to assemble all KCL constraints into a single linear system and \emph{directly} project onto its solution space using the Moore-Penrose pseudoinverse \cite{Penrose1955AGI}. 

\paragraph{Constructing the System} \(A\,\mathbf{y} + \mathbf{b} = 0\).
Recall eqs.~\eqref{eq:KCL1}--\eqref{eq:KCL2}, which describe KCL at each bus~\(i\).
Each of these can be written as a dot-product constraint:
\[
\langle\mathbf{a}^{P,i} ,\mathbf{y} \rangle \;+\; P_{\text{net}}(i) \;=\; 0,
\qquad
\langle \mathbf{a}^{Q,i} ,\mathbf{y} \rangle \;+\; Q_{\text{net}}(i) \;=\; 0,
\]
where \(\mathbf{a}^{P,i}\) and \(\mathbf{a}^{Q,i}\) are the normal vectors defined earlier. 

The main idea is now to stack all these constraints into a linear system
\[
A\,\mathbf{y} \;+\; \mathbf{b} \;=\; \mathbf{0},
\]
where \(A\in \mathbb{R}^{m\times d}\) and \(\mathbf{b}\in \mathbb{R}^m\), with $d \;=\; 4\,|E|,
m \;=\; 2\,N$. Concretely,
\[
A
~\defeq~
\begin{pmatrix}
(\mathbf{a}^{P,1})^\top \\[-3pt]
(\mathbf{a}^{P,2})^\top \\[-3pt]
\vdots            \\[-3pt]
(\mathbf{a}^{P,N})^\top \\[-1pt]
(\mathbf{a}^{Q,1})^\top \\[-3pt]
(\mathbf{a}^{Q,2})^\top \\[-3pt]
\vdots                  \\[-3pt]
(\mathbf{a}^{Q,N})^\top
\end{pmatrix}
\qquad\text{and}\qquad
\mathbf{b}
~\defeq~
\begin{pmatrix}
P_{\text{net}}(1) \\[-3pt]
P_{\text{net}}(2) \\[-3pt]
\vdots            \\[-3pt]
P_{\text{net}}(N) \\[-1pt]
Q_{\text{net}}(1) \\[-3pt]
Q_{\text{net}}(2) \\[-3pt]
\vdots            \\[-3pt]
Q_{\text{net}}(N)
\end{pmatrix}.
\]
 So that, solving \(A\,\mathbf{y}+\mathbf{b}=\mathbf{0}\) enforces \emph{all} KCL equations at once.

\medskip

\paragraph{Moore-Penrose Approach \cite{moore_proj}.}
An \emph{orthogonal} (least-squares) projection onto the affine space of admissible solutions 
\(\{\mathbf{z}\in\mathbb{R}^d : A\,\mathbf{z}+\mathbf{b} = \mathbf{0}\}\)
is given by
\[
\widetilde{\mathbf{y}}
~=~
\mathbf{y}
~-~
A^\dagger\Bigl(A\,\mathbf{y} + \mathbf{b}\Bigr),
\]
where \(A^\dagger \in \mathbb{R}^{d\times m}\)  is the Moore–Penrose inverse of \(A\).  By definition, if we let
\[
A \;=\; U \,\Sigma\, V^\top
\]
be the singular-value decomposition (SVD) of \(A\) (where \(U\) and \(V\) are orthonormal, and \(\Sigma\) is diagonal in the nonzero singular values), then
\[
A^\dagger
~\defeq~
V \,\Sigma^\dagger \,U^\top,
\]
with \(\Sigma^\dagger\) being the reciprocal of all nonzero singular values (and zero in any null dimensions).


\paragraph{Optimization View.}
This global projection via pseudoinverse solves the problem
\[
\min_{\widetilde{\mathbf{y}}}~
\|\widetilde{\mathbf{y}} - \mathbf{y}\|^2
\quad
\text{subject to}
\quad
A\,\widetilde{\mathbf{y}} + \mathbf{b} = 0,
\]
i.e.\ it is the \emph{closest point in Euclidean distance} to the original prediction \(\mathbf{y}\) that satisfies \emph{all} KCL constraints.  Therefore, the Moore–Penrose projection is the \emph{optimal} way to solve the KCL equations system, in the least-squares sense, starting from prediction $\mathbf{y}$. This also spares the choice of the variable ordering $\sigma$ above, whose optimization is a hard combinatorial problem.

\paragraph{Algorithmic Steps.}
One can thus enforce KCL as follows:

\begin{algorithmic}[1]
\REQUIRE $\mathbf{y} \in \mathbb{R}^d$ \quad (\textit{initial predicted flows})
\REQUIRE $A \in \mathbb{R}^{m \times d}$, \; $\mathbf{b} \in \mathbb{R}^m$, $A^{\dagger} \in \mathbb{R}^{d \times m}$ \quad (\textit{KCL system})
\STATE \(\mathbf{r} \,\leftarrow\, A\,\mathbf{y} + \mathbf{b}\) \quad  // KCL residual
\STATE \(\widetilde{\mathbf{y}} \,\leftarrow\, \mathbf{y} - A^\dagger \,\mathbf{r}\)
\RETURN \(\widetilde{\mathbf{y}}\)
\end{algorithmic}

\noindent
As above, because \(A\) and \(A^{\dagger}\) only depend on the \emph{topology} of the power grid, they can be assembled \emph{once} for a fixed network structure. In addition, this projection can be implemented in practice as two fully connected linear layers, with a residual connection for the second one, which allows for full back-propagation and thus training of the full architecture, e.g. via stochastic gradient descent.\\

\noindent
\textbf{Physically Consistent Predictions.} Integration of the global Moore-Penrose pseudoinverse projection step described above, into the model's architecture, as a last layer, ensures that the final predicted flows satisfy the power conservation principle. In contrast, as discussed in Sec. \ref{sec:lit-rev}, most methods in the literature merely direct the model towards lower KCL violations by incorporating them into the loss function, thereby offering no assurances regarding physical consistency.

\section{Experiments}
\label{sec:exp}
In this section, we systematically evaluate our proposed approach with two primary objectives. First, we benchmark the model against state-of-the-art methods on standard IEEE test cases under both nominal and contingency (N-1) conditions, focusing on predictive accuracy and adherence to physical constraints. Second, we perform an ablation study to assess specifically the impact of the final projection layer on performance and physical feasibility.
\subsection{Datasets}
To conduct our experiments, we employ two widely used benchmark datasets from the power engineering community and have also become standard benchmarks in recent deep-learning research: the IEEE 14-bus and IEEE 118-bus test cases. Each dataset comprises a single grid instance with nominal values assigned to every bus, containing 14 and 118 buses, respectively. To adapt these datasets for a machine learning framework, we augment the available data by generating multiple independent and identically distributed (iid) instances by sampling near the nominal values.
New data instances are generated for all buses by sampling from a normal distribution centered at their nominal values with a small variance. Specifically, the active power, voltage magnitude, and reactive power are sampled as follows:
$
P \sim \mathcal{N}\bigl(\bar{P},\, 0.01\bigr),\quad V \sim \mathcal{N}\bigl(\bar{V},\, 0.01\bigr),\quad Q \sim \mathcal{N}\bigl(\bar{Q},\, 0.01\bigr),
$
where the bar over the variable denotes the nominal value of the corresponding variable. After generating these perturbed grid parameters, we obtain the ground truth for the targeted variables using the Newton-Raphson method mentioned above.

\subsection{Implementation Details}

In this study, we simulated $20000$ distinct operational scenarios for power grids, capturing a comprehensive range of realistic conditions representative of real-time grid operations. Additionally, to assess model robustness under contingency conditions, we generated N-1 contingency scenarios by removing a single transmission line at random, while excluding the one that is directly connected to the slack bus.
In particular, these N-1 cases were excluded from the training set, ensuring an unbiased evaluation of the robustness of the models to realistic grid perturbations. We trained our models using the AdamW optimizer \cite{loshchilov2017decoupled}. The learning rate was equal to $10^{-3}$, while Xavier normal initialization \cite{glorot2010understanding} was applied to the linear layers of the proposed network. All training and evaluation processes were performed on a server with a single NVIDIA GeForce RTX $4080$ SUPER GPU.

\paragraph{Metrics:} To assess the accuracy of our models, we employ the Mean Squared Error (MSE) with respect to the ground-truth as a primary metric. Indeed, this method is the gold standard in the literature. Additionally, to ensure physical feasibility, we quantify the average KCL satisfaction for each grid. Mathematically, let \(N\) denote the total number of buses in the network and $P_{\text{net}}(i)$ and $P_{\text{calc}}(i)$ denote respectively the net injected and calculated power at bus $i$, as in \ref{sub:proj}. The global energy conservation loss for active power is then defined as the mean squared mismatch between the measured and calculated injections:
\[
\mathcal{L}_P = \frac{1}{N}\sum_{i=1}^{N} \left(P_{\text{net}}(i) + P_{\text{calc}}(i)\right)^2,
\]
and similarly for reactive power:
\[
\mathcal{L}_Q = \frac{1}{N}\sum_{i=1}^{N} \left(Q_{\text{net}}(i) + Q_{\text{calc}}(i)\right)^2.
\]
Finally, the overall physics-informed loss, which quantifies the degree of Kirchhoff’s Current Law (KCL) satisfaction, is given by:
\[
\mathcal{L}_{\text{KCL}} = \frac{1}{2}\left(\mathcal{L}_P + \mathcal{L}_Q\right).
\]

\subsection{Comparison with other models}
We compare our model against three well-established models from the literature \cite{donon2020leap,kody2022modeling,lin2024powerflownet}. Note, that since the classical Newton-Raphson solver already generates the ground-truth labels for our datasets, using it as a benchmark would be redundant, since by definition it will show \textit{zero error}. However, graph-based approaches have showcased much faster performance in previous works \cite{lin2024powerflownet}, which also highlights the importance of these surrogate models. The selection of these models was motivated by their diverse approaches to integrating physics-informed knowledge: one model omits such integration entirely, another incorporates partial domain knowledge through its architectural design, and the third enforces a soft physics constraint via its loss function. Since the power flow prediction problem has not yet been uniformly defined, each architecture addresses the task within its own formal framework (as discussed in Section \ref{sec:lit-rev}). Therefore, since PowerFlow predicts node features, we compute the average squared mismatch using the equations in \ref{eq:KCL-ij}.

\begin{table}[h]
  \centering
  \setlength{\tabcolsep}{16pt}
  \renewcommand{\arraystretch}{1.2}
  \begin{tabular}{l l c c}
    \toprule
    \textbf{Dataset} & \textbf{Model} & \textbf{MSE} & \textbf{KCL Violation}  \\
    \midrule
    \multirow{4}{*}{\textbf{IEEE118}} 
      & LeapNet  & 0.591 \scriptsize{(0.001)} & 1.169 \scriptsize{(0.019)} \\
      & KodyNet  & 0.298 \scriptsize{(0.004)} & 0.359 \scriptsize{(0.002)} \\
      & PowerFlowNet & 0.642 \scriptsize{(0.003)} & 0.327 \scriptsize{(0.001)} \\
      & KCLNet  & \textbf{0.273 \scriptsize{(0.002)}} & \textbf{0.000 \scriptsize{(0.000)}} \\
    \midrule
    \multirow{4}{*}{\textbf{IEEE14}}
      & LeapNet  & 0.439 \scriptsize{(0.001)} & 1.664 \scriptsize{(0.148)} \\
      & KodyNet  & \textbf{0.169 \scriptsize{(0.001)}} & 1.288 \scriptsize{(0.093)} \\
      & PowerFlowNet & 0.746 \scriptsize{(0.028)} & 1.933 \scriptsize{(0.042)} \\
      & KCLNet & 0.381 \scriptsize{(0.005)} & \textbf{0.000 \scriptsize{(0.000)}} \\
    \bottomrule
  \end{tabular}
  \caption{Performance results in N-case: We report the average of 10 runs of the MSE and KCL violation for each model on the IEEE 14-bus and IEEE 118-bus datasets under nominal operating conditions. For each metric, the corresponding standard deviation (std) is provided in parentheses. The model with the best (lowest) MSE is highlighted in bold. Values reported as 0.000 indicate numerical errors on the order of 1e-4, with all results rounded to three decimal places.}
  \label{tab:results-N}
\end{table}

\subsection{N-case Analysis} 
In the nominal operating scenario (N-case), we evaluated model performance on the IEEE 14-bus and IEEE 118-bus test cases. As shown in Table \ref{tab:results-N}, the observed MSEs ranged from 0.169 to 0.746. Although KodyNet reported a lower MSE on the IEEE 14-bus case (0.169), it also exhibited a relatively higher mismatch in Kirchhoff’s Current Law (KCL) (1.288). In contrast, KCLNet maintained zero KCL violation, with MSE values of 0.381 on IEEE14 and 0.273 on IEEE118. 
We observe that the predictions obtained using the larger IEEE118 network generally outperform those from smaller networks. This can be attributed to the inherently local nature of power flow prediction, whereby localized errors have a reduced impact on the overall grid performance in larger systems compared to more confined network models.

\subsection{N-1 Case Analysis}

Under N-1 contingency conditions, where a single transmission line is removed to simulate grid perturbations, the overall MSE increased across all models, as detailed in Table \ref{tab:results-N-1}. Notably, KCLNet continued to enforce complete compliance with KCL (zero violation), while recording MSE values of 0.533 on IEEE14 and 0.2768 on IEEE118. Furthermore, it is evident that within the larger IEEE118 dataset, the MSE of the proposed architecture is close to that observed in the complete network (N-case). A natural interpretation is that the removal of a single transmission line predominantly affects only a localized region of the grid, rather than inducing widespread, global changes in performance. The other models, despite achieving comparable MSEs, displayed non-negligible KCL violations. These observations suggest that embedding physical constraints into the modeling framework can contribute to maintaining feasibility under both nominal and perturbed grid conditions, although further analysis is required to fully elucidate the trade-offs involved.

\begin{table}[h]
  \centering
  \setlength{\tabcolsep}{16pt}
  \renewcommand{\arraystretch}{1.2}
  \begin{tabular}{l l c c}
    \toprule
    \textbf{Dataset} & \textbf{Model} & \textbf{MSE} & \textbf{KCL Violation}  \\
    \midrule
    \multirow{4}{*}{\textbf{IEEE118}} 
      & LeapNet  & 0.643 \scriptsize{(0.002)} & 1.149 \scriptsize{(0.061)} \\
      & KodyNet  & 0.591 \scriptsize{(0.003)} & 0.361 \scriptsize{(0.007)} \\
      & PowerFlowNet & 0.981 \scriptsize{(0.001)} & 0.845 \scriptsize{(0.001)} \\
      & KCLNet & \textbf{0.2768 \scriptsize{(0.000)}} & \textbf{0.000 \scriptsize{(0.000)}}  \\
    \midrule
    \multirow{4}{*}{\textbf{IEEE14}}
      & LeapNet & 0.452 \scriptsize{(0.004)} & 1.632 \scriptsize{(0.109)} \\
      & KodyNet & \textbf{0.199 \scriptsize{(0.001)}} & 1.292 \scriptsize{(0.130)} \\
      & PowerFlowNet & 0.845 \scriptsize{(0.023)} & 1.629 \scriptsize{(0.040)} \\
      & KCLNet & 0.533 \scriptsize{(0.003)} & \textbf{0.000 \scriptsize{(0.000)}} \\
    \bottomrule
  \end{tabular}
  \caption{Performance results in N-1 case: We report the average of 10 runs of the Mean Squared Error MSE and KCL violation for each model on the IEEE 14-bus and IEEE 118-bus datasets under nominal operating conditions. For each metric, the corresponding standard deviation (std) is provided in parentheses. The model with the best (lowest) MSE is highlighted in bold. Values reported as 0.000 indicate numerical errors on the order of 1e-4, with all results rounded to three decimal places.}
  \label{tab:results-N-1}
\end{table}

\subsection{Ablation Study}
In this section, we assess the impact of the final projection layer of our KCLNet model. Specifically, we remove the KCL projection layer to relax the hard physical constraints during training, thereby expanding the loss space. We observe that removing the final projection layer generally allows the network to achieve a lower MSE, which is expected as the loss space becomes less constrained by the hard projection (Table \ref{tab:ablation}). For example, under the N-case scenario on IEEE14, the MSE drops from 0.381 (with projection) to 0.353, and similarly, in the N-1 case the MSE reduces from 0.533 to 0.492. Although for IEEE118 under normal conditions the MSE shows a slight increase (from 0.273 to 0.296), the overall trend suggests that the relaxation permits better optimization of the primary objective. Notably, the corresponding KCL violations are relatively low (ranging from 0.039 to 0.047), slightly below those seen in the other models. This indicates that even without the final projection layer enforcing strict physical constraints, the model still achieves a reasonable level of physical feasibility while benefiting from a more flexible loss landscape.

\begin{table}[h]
  \centering
  \setlength{\tabcolsep}{7.2pt}
  \renewcommand{\arraystretch}{1.2}
  \begin{tabular}{l c c c c}
    \toprule
    \multirow{2}{*}{\textbf{Dataset}} & \multicolumn{2}{c}{\textbf{N Case}} & \multicolumn{2}{c}{\textbf{N-1 Case}} \\
    \cmidrule(lr){2-3} \cmidrule(lr){4-5}
     & \textbf{MSE} & \textbf{KCL Violation} & \textbf{MSE} & \textbf{KCL Violation}  \\
    \midrule
    \textbf{IEEE118} & 0.296 \scriptsize{(0.002)} & 0.039 \scriptsize{(0.001)} & 0.272 \scriptsize{(0.000)} & 0.047 \scriptsize{(0.002)} \\
    \textbf{IEEE14}  & 0.353 \scriptsize{(0.001)} & 0.043 \scriptsize{(0.001)} & 0.492 \scriptsize{(0.001)} & 0.046 \scriptsize{(0.001)} \\
    \bottomrule
  \end{tabular}
  \caption{Performance of KCLNet without the final projection layer under nominal (N case) and contingency (N-1 case) conditions. We report the mean and standard deviation in parentheses, calculated over 10 runs.}
  \label{tab:ablation}
\end{table}

\section{Conclusion}
\label{sec:conclusion}

In this work, we introduced a physics-informed machine learning approach for power flow prediction that integrates Kirchhoff’s Current Law (KCL) as a hard constraint via hyperplane projections. Our proposed model, \emph{KCLNet}, consists of a Graph Neural Network architecture followed by a physics-informed projection layer, ensuring every prediction remains physically plausible under both nominal and contingency (\(N{-}1\)) operating conditions. Experimental evaluations on the IEEE 14-bus and IEEE 118-bus test cases show that KCLNet, while maintaining competitive Mean Squared Error (MSE) values on smaller grids, achieves the lowest MSE on the larger IEEE 118-bus system. Meanwhile, it enforces KCL with zero numerical violation—an outcome contrasting with other state-of-the-art methods that, despite occasionally achieving lower MSEs on smaller test grids, exhibit non-negligible KCL deviations.

These results indicate that directly embedding physical constraints into the model architecture not only enhances reliability but also contributes to operational safety in real-world grid scenarios. Notably, the scalability and strong performance on the larger IEEE 118-bus system suggest that KCLNet is well-suited for more extensive networks in practice.
\paragraph{Future Extensions.}
Beyond our current scope, we next propose two directions that could further expand the applicability and impact of KCLNet,
\begin{itemize}
    \item \textbf{Nonlinear Network Equations and Diffusion-Based Projections:} 
    When the steady state of a physical system is governed by nonlinear equations, one could explore deterministic diffusion or similar iterative methods to approximate the projection on the nonlinear constraint manifold. \\
    
    \item \textbf{Application to Various Physical Networks Scenarios:} 
    The analogy between electrical flows and other resource flows is especially relevant in contemporary \emph{smart grid} and \emph{smart city} applications, where Kirchhoff-like conservation principles arise naturally. For instance, \cite{9357037} demonstrates a traffic management system guided by network “circuit” theorems akin to KCL. Extending KCLNet to such contexts promises physically consistent predictions  in a wide range of complex, networked systems.
\end{itemize}

\section*{Acknowledgments} This research is supported by FNR Luxembourg INTER/FNRS/20/15\-077233/ Scaling Up and CREOS S.A. We would like to thank Yves Reckinger and Robert Graglia for their valuable feedback.

\bibliographystyle{splncs04} 
\bibliography{biblio} 

\end{document}